\pdfoutput=1
\documentclass[11pt]{article}

\usepackage[]{acl}
\usepackage{times}
\usepackage{latexsym}

\usepackage[T1]{fontenc}
\usepackage[utf8]{inputenc}

\usepackage{microtype}

\usepackage{inconsolata}

\usepackage{graphicx}
\usepackage{amsmath}
\usepackage{amssymb}
\usepackage{algorithm}
\usepackage{algpseudocode}
\usepackage{booktabs}
\usepackage{multirow}

\title{OpenWebVoyager: Building Multimodal Web Agents via Iterative Real-World Exploration, Feedback and Optimization}

\author{Hongliang He$^{1,3}$\thanks{Work done during the internship at Tencent AI Lab.}, 
        Wenlin Yao$^{2}$\thanks{Work done while at Tencent AI Lab.}, 
        Kaixin Ma$^{2}$, 
        Wenhao Yu$^{2}$, 
        \textbf{Hongming Zhang}$^{2}$, \\
        \textbf{Tianqing Fang}$^{2}$,
        \textbf{Zhenzhong Lan}$^{3}$,
        \textbf{Dong Yu}$^{2}$ \\
        $^{1}$Zhejiang University,
        $^{2}$Tencent AI Lab (Seattle), 
        $^{3}$Westlake University \\
    }

\begin{document}
\maketitle
\begin{abstract}
% The advancement of large language models has laid the groundwork for building autonomous agents for complex tasks such as web navigation. 
The rapid development of large language and multimodal models has sparked significant interest in using proprietary models, such as GPT-4o, to develop autonomous agents capable of handling real-world scenarios like web navigation.
Although recent open-source efforts have tried to equip agents with the ability to explore environments and continuously improve over time, they are building text-only agents in synthetic environments where the reward signals are clearly defined. Such agents struggle to generalize to realistic settings that require multimodal perception abilities and lack ground-truth signals.
% Recent efforts have also tried to equip the agent with the ability to explore environments and continuously improve over time. However, existing works only focused on building text-only agents in synthetic environments where the reward signals are clearly defined. Such agents can hardly generalize to realistic settings that require multimodal perception ability and provide no ground-truth signal.
In this paper, we introduce an open-source framework designed to facilitate the development of multimodal web agent that can autonomously conduct real-world exploration and improve itself. 
% We first compile a set of high-quality trajectories collected using a state-of-the-art agent to train our agent and equip it with basic abilities. 
We first train the base model with imitation learning to gain the basic abilities.
%web task queries from existing datasets, human and automatic synthesis. The agent first conducts imitation learning from a small set of web navigation trajectories collected using a state-of-the-art agent built upon GPT-4o.
We then let the agent explore the open web and collect feedback on its trajectories. After that, it further improves its policy by learning from well-performing trajectories judged by another general-purpose model. 
This exploration-feedback-optimization cycle can continue for several iterations. 
Experimental results show that our web agent successfully improves itself after each iteration, demonstrating strong performance across multiple test sets.\footnote{Code and data will be released at \url{https://github.com/MinorJerry/OpenWebVoyager}. Contact: \texttt{hehongliang@westlake.edu.cn}}
\end{abstract}

\section{Introduction}
%Developing an autonomous Web Agent capable of dealing with complex real-world web environments and completing web navigation tasks has been a challenging goal within the AI community.
Developing autonomous agents that can complete complex tasks such as web navigation has been a significant challenge for the AI community \cite{zhou2023webarena,gur2023real,deng2024mind2web,koh2024visualwebarena}. Recent advancements of large language and multimodal models such as Claude \citep{anthropic2024introducing} and GPT-4o \citep{openai2024hello} have made it possible to build such agents via prompt engineering \cite{he2024webvoyager,zheng2024gpt4vision,ma2023laser}. However, these agents struggle to improve further due to their reliance on closed-source models. Another line of work has explored alternative ways to build agents by starting off with weaker open-source models and gradually improving model performance by iteratively exploring the environment, collecting feedback signals, and updating the policy model \citep{xi2024agentgym, putta2024agent, patel2024large}. 
%Developing an autonomous web agent that can effectively navigate complex, real-world web environments and successfully complete diverse web-based tasks has been a significant challenge for the AI community.
%The impressive imitation and reasoning capabilities of Large Language Models (LLMs) imbue them with the potential to construct generalist web agents 
%The advanced language understanding, generation, and reasoning capabilities of Large Language Models (LLMs) make them promising candidates for developing versatile web agents
%\citep{gur2023real, zhou2023webarena}. 
%To cope with complex web environments, as well as the lack of training data, previous works \citep{xi2024agentgym, putta2024agent, patel2024large} have explored ways to enhance web agents and exceed the base performance through learning via exploration. 
However, existing studies have only focused on building text-only agents in synthetic environments \citep{song2024trial}. The synthetic environments provide the benefit of well-defined reward signals, allowing the agents to effectively differentiate the quality of the trajectories and learn accordingly. However, synthetic environments fail to capture the complexity of real-world scenarios, leading to potential generalization issues when applied to real-world tasks. Moreover, real-world environments often do not have built-in reward signals, which poses another challenge in agent's learning and improvement process~\cite{he2024webvoyager}. Additionally, real-world webpages are designed based on human visual preference, ignoring the visual inputs can cause significant information loss that impacts the agent's performance.

% Enabling web agents powered by open-source LMMs to continuously learn and improve themselves through interactions with websites, similar to how humans learn by exploring the web, is important to achieve self-evolving. Combining visual understanding of website layouts with natural language understanding of text content is the key to empowering open-source web agents with human-like self-exploratory learning capabilities.
% However, there is a lack of exploration of how Large Multimodal Models (LMMs) based web agents can perform self-improvements. Current LMM-based web agents no longer rely solely on HTML for reasoning and navigation; instead, they utilize multimodal information \citep{zhou2023webarena, koh2024visualwebarena, zheng2024gpt, he2024webvoyager}. This paradigm allows them to handle multimodal web tasks, aligns with user experience, and effectively leverages browser rendering capabilities, thus becoming mainstream. Existing studies \citep{he2024webvoyager, koh2024visualwebarena} have explored building LMM-based Web Agents on powerful closed-source models like Claude \citep{anthropic2024introducing} and GPT-4o \citep{openai2024hello}, achieving acceptable results, yet challenges remain in performing Supervised Fine-Tuning (SFT) and Self-improvements on open-source multimodal models for real-world online web environments. 

\begin{figure*}[t]
\centering
\includegraphics[width=0.85\linewidth]{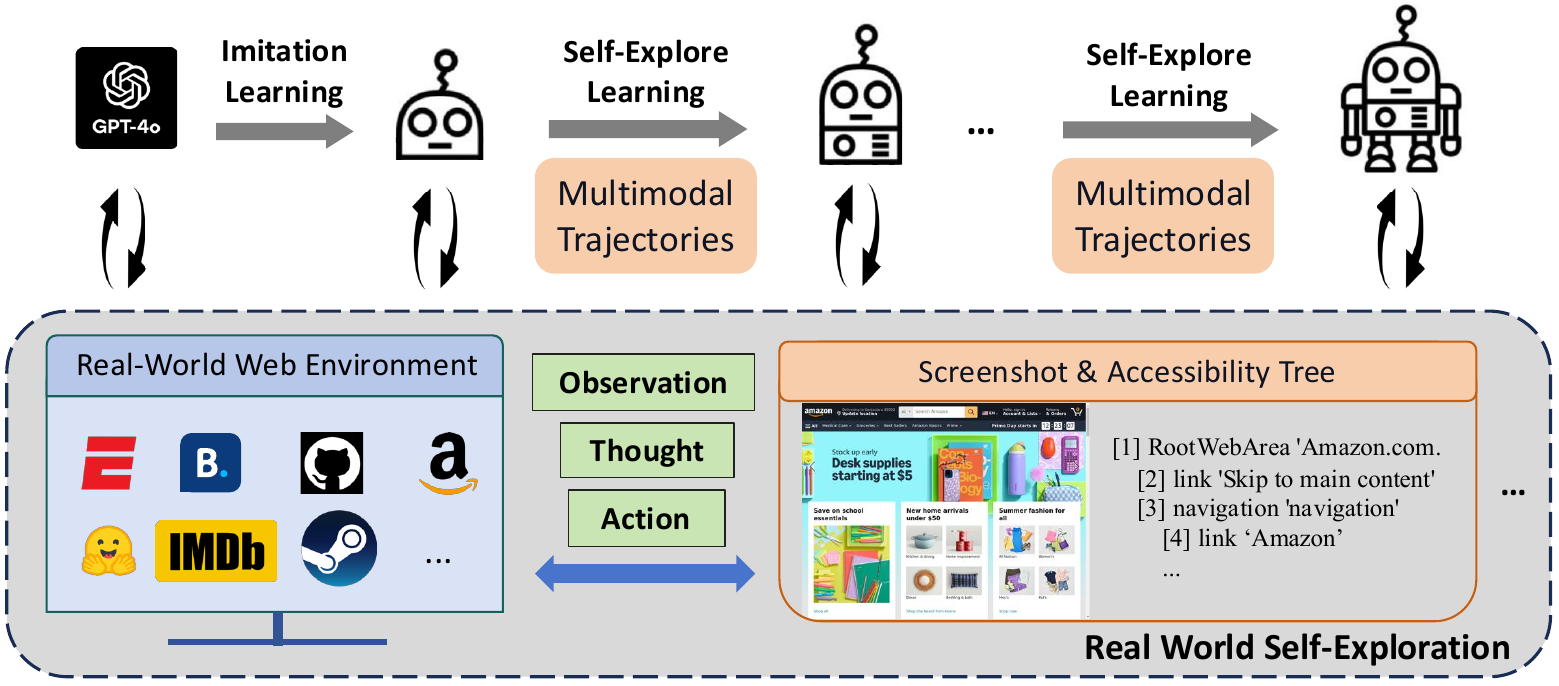}
\caption{The overall process of OpenWebVoyager, including the Imitation Learning phase and the exploration-feedback-optimization cycles. The agent learns basic multimodal web navigation skills through Imitation Learning and continues to explore real-world web environments. GPT-4o provides feedback on explored multimodal trajectories, leaving successful trajectories for the agent to improve.}%The agent learns basic multimodal web navigation skills through Imitation Learning and further explores and improves in the exploration-feedback-improve cycles. }
\label{fig:full_process}
\vspace{-0.05in}
\end{figure*}

To address above limitations and explore open-source models in real-world settings, we propose OpenWebVoyager, an open-source framework for building multimodal web agents via iterative real-world exploration, feedback and optimization.
We show that OpenWebVoyager can learn to perform real-world web navigation tasks through an initial \textit{imitation learning} (IL) phase followed by multiple exploration-feedback-optimization cycles. To do so, we start by compiling a diverse set of web task queries and collecting corresponding agent trajectories using a state-of-the-art multimodal agent WebVoyager \citep{he2024webvoyager} based on GPT-4o, which we refer to as WebVoyager-4o. During the imitation learning phase, we train OpenWebVoyager on trajectories where WebVoyager-4o successfully completes the task to teach the agent basic skills to perform web navigation. %Trajectories gathered by GPT-4o are usually not optimal but have more reflection and error correction information. 
%and leverage the WebVoyager framework \citep{he2024webvoyager} to generate multimodal trajectories. GPT-4o \citep{openai2024hello} serves as a proficient guide in web navigation, and the gathered trajectories enable our agent to grasp basic web knowledge and operational logic. Trajectories gathered by GPT-4o are usually not optimal but have more reflection and error correction information. 
Subsequently, within the exploration-feedback-optimization cycle, we continue to synthesize new web tasks, allowing our agent to explore and gather more trajectories. During this stage, we follow \citet{he2024webvoyager} and leverage GPT-4o to automatically evaluate the correctness of the trajectories produced by OpenWebVoyager. After gathering feedbacks, we retain successful trajectories and merge them with the data from imitation learning phase %into the training data pool 
to conduct the next round of training to improve OpenWebVoyager. 
The improved agent is then used to sample new trajectories in the next iteration.  
%We suggest employing GPT-4o to provide feedback because only close-source multimodal models and human judgment are adept for complex multimodal navigation trajectories. 
This streamlined and effective design frees us from the limitations and obsolescence of manually collected trajectories, relying more on GPT-4o's supervision, thus bringing the feasibility of continuous optimization.

In our experiments, we employ idefics2-8b-instruct \citep{laurençon2024mattersbuildingvisionlanguagemodels} as our backbone model and 
select 48 common websites from the WebVoyager and Mind2Web datasets \cite{deng2024mind2web} to gather trajectories. The overall process includes one imitation learning phase and three exploration-feedback-optimization cycles. For each phase, we leverage self-instruct \citep{wang2022self} to generate new web queries. %(Hongliang double check this)
%At each phase, we assess the Task Success Rate using the WebVoyager and Mind2Web test datasets, employing GPT-4o for automated evaluation. 
We assess the agent's performance using the Task Success Rate on the WebVoyager and Mind2Web test sets.
Results indicate a gradual increase in task success rate across the four phases on the WebVoyager test set from 19.9\% to 25.8\% and on the Mind2Web cross task set from 6.3\% to 19.6\%, demonstrating the potential for iterative optimization in multimodal web agents. Additionally, a slight improvement is observed on the Mind2Web cross-web (unseen web) set from 6.6\% to 10.4\%, suggesting that the exploration-feedback-optimization cycle can, to some extent, generalize to unseen websites.

\section{Related Work}
\subsection{Multimodal Web Agents}
%Recently, research on multimodal (usually vision-language) web agents has been increasing. 
Recently, there has been a growing interest in building multimodal web agents, particularly those that combine visual and textual understanding capabilities.
Unlike traditional HTML-dependent LLM-based agents \citep{lutz2024wilbur, zhou2023webarena, gur2023real, nakano2021webgpt, ma2023laser}, Large Multimodal Model (LMM)-based agents can perform a wider variety of web tasks and adapt to more complex web environments. The main difference lies in the observation space. To acquire multimodal input signals, SeeAct \citep{zheng2024gpt} focuses on annotating images of web pages using bounding boxes and index labels of candidate web elements. WebVoyager \citep{he2024webvoyager} and VisualWebArena \citep{koh2024visualwebarena} both use a JavaScript tool to extract web elements and annotate them on screenshots in a Set-of-Mark \citep{yang2023set} format. DUAL-VCR \citep{kil2024dual} contextualizes each web element with its neighbors in the screenshot. SCAFFOLD \citep{lei2024scaffolding} introduces dot matrices and coordinates on images to enhance visual grounding. Most of the aforementioned multimodal web agents rely on prompting closed-source multimodal models such as GPT-4V \citep{openai2023gpt4}, Claude \citep{anthropic2024introducing}, and Gemini \citep{team2023gemini}. 
%Prompting closed-source multimodal models such as GPT-4V \citep{openai2023gpt4}, Claude \citep{anthropic2024introducing}, and Gemini \citep{team2023gemini} to build LMM-based web agents is widely adopted, due to 
These models' strong visual grounding and understanding capabilities enable them to correctly interpret webpage screenshots and engage in proper planning using paradigms like ReAct \citep{yao2022react} or Chain-of-Thought \citep{wei2022chain}. 
%On the other hand, open-source models often suffer from significant performance degradation. SeeAct and VisualWebArena demonstrate that fine-
%tuning has minimal effects on 
While some previous works attempted to leverage open-source vision-language models to build web agents \cite{zheng2024gpt,koh2024visualwebarena}, they found that
models such as BLIP-2-T5 \citep{jian2024bootstrapping}, LLaVA \citep{liu2024visual}, and Idefics \citep{laurençon2023obelicsopenwebscalefiltered} can hardly achieve satisfactory performance. 
The main reason is that the pretraining of those open-source vision-language models mostly focuses on aligning image-text features and visual question answering instead of image-text interleaved agent trajectories. 
In this work, we propose an agent built upon an open-source model that can automatically collect trajectories to continuously improve itself, leading to salient gains in performance. %which primarily focuses on VQA training and lacks proficiency in handling long trajectories. Another challenge is the lack of training data, as existing datasets such as Multimodal-Mind2Web \citep{zheng2024gpt}, VisualWebBench \citep{liu2024visualwebbench}, and OmniAct \citep{kapoor2024omniact} 
%are difficult to unify into a cohesive training dataset due to differences in their structure, format, and annotations.

\subsection{Self-Improving Web Agents}
Researchers also have attempted to boost agents and adapt them to complex environments through self-improvement. AgentGYM \citep{xi2024agentgym} proposes a framework that unifies a wide range of environments for real-time exploration and evolution of LLM-based agents. AgentQ \citep{putta2024agent} integrates Monte Carlo Tree Search (MCTS) and Direct Preference Optimization (DPO; \citealp{rafailov2024direct}) algorithms to iteratively update the policy of LLM-based web agents based on successful and failed web trajectories. \citet{patel2024large} suggests improvement by utilizing web agents to collect and filter in-domain trajectories, plus out-of-domain tasks along with hypothetical solution trajectories. However, there is still a lack of exploration on how to leverage multimodal web signals to achieve self-improvement. We aim to enable multimodal web agents to adapt to complex and dynamic online environments, enhancing their generality and ability to operate across numerous online websites.

\begin{figure*}[t]
\centering
\includegraphics[width=0.85\linewidth]{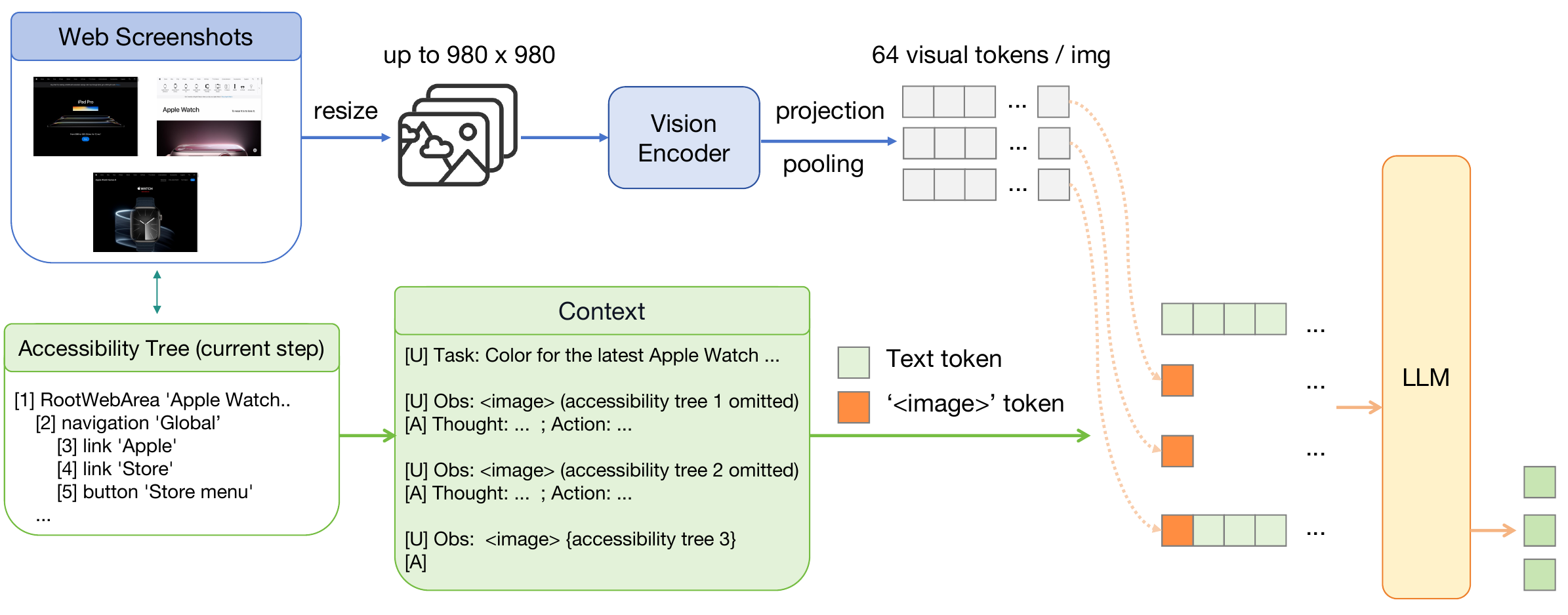}
\caption{
%Using Idefics2 structure to train a multimodal web agent.
The model architecture of our multimodal web agent.
We use the most recent 3 web screenshots to demonstrate the page changes after performing web actions and label the web elements in the accessibility tree to facilitate the agent in selection and response. Considering the limitation of sequence length and to avoid confusion, we only retain the most recent accessibility tree.}
\label{fig:idefics2}
\vspace{-0.05in}
\end{figure*}

\section{Method}
%With the online web environment growing in complexity and dynamism, training practical Large Web Agents has become increasingly challenging, primarily due to: 
% The complexity of the unseen and ever-evolving web environment has made training practical web agents particularly challenging, primarily due to two reasons.
% 1) Existing web navigation trajectories are mostly collected in a static environment and will not iteratively update; 2) these annotated trajectories typically represent the shortest path to a goal, lacking the exploration needed for agents to learn from both successes and failures. 
In this section, we introduce OpenWebVoyager, an innovative web agent that outlines a path of iterative optimization for LMM-based Web Agents to handle intricate online web tasks. Firstly, we enable the agent to learn web navigation trajectories of WebVoyager-4o in the first stage to gain basic web knowledge and navigation skills, namely Imitation Learning (IL). Subsequently, the agent iteratively explores and improves with the feedback from GPT-4o.

\subsection{Task Formulation}
In the web browsing environment $\mathcal{E}$, consider the web navigation process as a Partially Observable Markov Decision Process (POMDP). The setup is defined by the tuple $(\mathcal{S}, \mathcal{O}, \mathcal{A}, \mathcal{T}, R)$, where $\mathcal{S}$ denotes the state space, $\mathcal{O}$ represents the observation space, and $\mathcal{A}$ is the action space. $\mathcal{T}$ is the deterministic transition function that performs web operations in the browser to promote the process. The reward $R$ in this environment is typically a sparse signal indicating success or failure, with values of 1 or 0, respectively.

Given a task query $q$ and its corresponding website $w$, we can initialize the web environment $\mathcal{E}$ by setting the state $s_1$ to this web page, and obtain the first step observation $o_1 \in \mathcal{O}$. In this work, we adopt the vision-language setting that the observation in each step will include an accessibility tree and a screenshot, i.e., $o_1 = (o^{a}_1, o^{s}_1)$.  Let $\theta$ represents the parameters of the Large Multimodal Models (LMMs). Following the ReAct paradigm, we derive thoughts and actions using LMMs: $(h_1, a_1) \sim \pi_{\theta}(\cdot |I, q, o_1) = \pi_{\theta}(\cdot | I, q, o^{a}_1, o^{s}_1)$, where $I$ denotes the system prompt, including answer formats, the introduction of web operations and some guidelines. The transition function $\mathcal{T}$ is then applied to parse the action and execute it on the web page, obtaining the next state $s_2$. Therefore, at time step $t$, we have: 
\begin{equation}
    (h_{t}, a_{t}) \sim \pi_{\theta}(\cdot | I, q, o^a_1, o^s_1, h_1, a_1,..., o^a_t, o^s_t)
\end{equation}
\begin{equation}
    s_{t+1} = \mathcal{T}(s_t, a_t; \mathcal{E}).
\end{equation}
The full trajectory can be represented as $\tau = (o^a_1, o^s_1, h_1, a_1, ..., o^a_T, o^s_T, h_T, a_T)$, where $T$ is the number of iterations in web navigation, i.e., the length of the trajectory.

\subsection{OpenWebVoyager Overview}
\paragraph{Environment}
We adopt the Selenium-based online web navigation environment provided by WebVoyager \citep{he2024webvoyager}. In contrast to WebVoyager, we do not employ the Set-of-Mark approach to mark elements on screenshots because open-source LMMs face significant visual grounding issues in identifying numerical tags on screenshots. We modify the observation of the web page to include the accessibility tree and its corresponding unmarked screenshot. Figure \ref{fig:observation_space} in Appendix \ref{sec:appendix-env-prompt} shows a specific example of the observation space.

\paragraph{Model and Learning}
We adopt Idefics2 \citep{laurenccon2024matters} as the backbone LMM for building OpenWebVoyager. Idefics2 is well-suited for our task as it incorporates interleaved image-text documents during training, boosting the model's multi-image reasoning and long-context comprehension capabilities. Additionally, Idefics2 supports encoding high-resolution images up to 980x980 pixels, which is necessary for preserving the fine-grained visual details on the webpage screenshots. %utilizes pooling to reduce the number of visual tokens, thereby improving efficiency. 
% It achieves state-of-the-art performance across various multimodal benchmarks.
%employs OBELICS \citep{laurençon2023obelicsopenwebscalefiltered}, 
%incorporates a wealth of interleaved image-text documents rather than solely relying on image-text pairs, which will be beneficial for text-rich web navigation tasks.
In Figure \ref{fig:idefics2}, we elaborate on how we adapt the Idefics2 architecture to build OpenWebVoyager. Similar to the messages fed into GPT-4o, we embed the <image> token at the corresponding position in the context, aligning it with the accessibility tree. The Idefics2-based agent will make a decision based on the observation containing multimodal information.
Figure \ref{fig:full_process} illustrates the full process of IL and exploration-feedback-optimization cycle: collecting trajectories for Imitation Learning via WebVoyager-4o, training the base agent, and then continuously exploring new trajectories. Based on feedback from GPT-4o, successful trajectories are leveraged for optimization.

\subsection{Web Task Queries Collection}
\paragraph{Queries for the Imitation Learning Phase} 
The IL phase is crucial as it forms the foundation for subsequent improvements. We aim to gather a diverse set of web tasks of varying difficulty, enabling GPT-4o to generate diverse trajectories.
We choose 48 popular websites, then select and synthesize the queries $\mathcal{Q}_{\text{IL}}$ from multiple perspectives before Imitation Learning. The details of $\mathcal{Q}_{\text{IL}}$ collection are shown in Appendix \ref{sec:appendix-datasets}.
%Specifically, we select and synthesize the following queries $\mathcal{Q}_{\text{IL}}$ from multiple perspectives before Imitation Learning: 

\paragraph{Queries for Real-World Exploration}
We continue to use the self-instruct \citep{wang2022self} approach to generate new queries that are similar but not duplicated based on existing queries. In each exploration-feedback-optimization cycle, we automatically generate 480 queries for 48 websites, with 10 queries for each website. The agent then conducts web exploration based on these tasks.

\subsection{Imitation Learning}
\label{section:IL_stage}

\paragraph{Trajectories Collection}
We utilize GPT-4o along with the WebVoyager paradigm \citep{he2024webvoyager} to generate web navigation trajectories corresponding to the above queries. The agent is named WebVoyager-4o and configured to receive observations consisting of the latest $k$ steps, including the accessibility trees and screenshots. 
i.e., for each $q_i \in \mathcal{Q}_{\text{IL}}$, $\tau_i \sim \pi_{\theta_g}(\tau | I, q_i)$, we clip the long context $c_t$ to avoid performance degeneration when $t>k$:
\begin{multline} \label{equation:c-clip}
    c_t^{\text{clip}} = (h_1,a_1,h_2,a_2,...,h_{t-k}, a_{t-k}, \\ o_{t-k+1}, h_{t-k+1}, a_{t-k+1}, ..., o_t),
\end{multline}
\begin{equation} \label{equation:gpt-4o-reasoning}
    (h_{t}, a_{t}) \sim \pi_{\theta_g}(\cdot | I, q, c_t^{\text{clip}}).
\end{equation}
It is worth noting that we preserve the thought and action of each step to maintain the full reasoning process without occupying excessive context. 
The collected trajectories fall into three pre-defined categories: \textbf{unfinished} (exceeding the maximum iteration of Navigation), \textbf{finished \& unsuccessful}, and \textbf{finished \& successful}. In this stage, to better distill knowledge from GPT-4o, we filter out unfinished trajectories, retaining only the other ones for training in Imitation Learning. Meanwhile, we resample the unfinished tasks once to improve the utilization of queries and reduce the problem of navigation failure due to sampling randomness.

\paragraph{Learning}
We adopt Idefics2 \citep{laurenccon2024matters} to learn trajectories collected through WebVoyager-4o. In Idefics2, screenshots are encoded as 64 visual tokens. %but the accessibility trees are typically long. 
However, the length of each accessibility tree is typically way longer than 64 tokens.
Considering the sequence length issue, we have to further truncate the context and the number of images, retaining the latest $k$ images while keeping only one accessibility tree of the current page. 
That is, we remove $k-1$ accessibility trees in Equation \ref{equation:c-clip}, 
\begin{multline}
    c_t^{\text{clip}'} = (h_1,a_1,...,h_{t-k},a_{t-k}, \\ o_{t-k+1}^s, h_{t-k+1}, a_{t-k+1}, ..., o_t^s, o_t^a).
\end{multline}
Let $D_{\text{IL}}$ represents the collected trajectories, and $\theta$ denote the parameters of the Idefics2 model. We aim to maximize the following objective function:
\begin{multline}\label{equation:J-bc}
    \mathcal{J}_{\text{IL}}(\theta) = \mathbb{E}_{(q,\tau) \sim D_{\text{IL}}} \sum_{t=1}^T \Big[\log \pi_\theta(a_t | q, c_t^{\text{clip}'}, h_t) \\ + \log \pi_\theta(h_t | q, c_t^{\text{clip}'})\Big],
\end{multline}
where the system prompt $I$ is no longer provided because of its considerable length. Through Imitation Learning, the agent has already learned the basic operation logic and response format, so there is no need for the system prompt.

\subsection{Iterative Optimization}
After the Imitation Learning phase, the trained agent $\pi_{\theta_b}$ will proceed to explore websites and undergo multiple cycles of exploration-feedback-optimization. We continue to generate more queries using self-instruct. Instead of relying on WebVoyager-4o to collect trajectories, the agent collects trajectories on its own. 
At each exploration-feedback-optimization cycle, we employ trajectory-level rejection sampling via GPT-4o to ensure quality trajectories. Let $Q_{\text{SI}}^j$ be the query set for $j$-th optimization, for every $q \in Q_{\text{SI}}^j$, we sample several trajectories from the model $\pi_{\theta_{j-1}}$, with GPT-4o acting as the Auto Evaluator, accepting only trajectories that GPT-4o deems as successfully navigated. We consider this auto evaluation method reliable because assessing the correctness of a trajectory is much easier than obtaining a correct trajectory. \citet{he2024webvoyager} also demonstrates a high level of evaluation consistency between GPT-4o and humans.

Let $D_{\text{SI}}^j$ represent the set of trajectories collected after rejection sampling in the $j$-th optimization. We mix the collected trajectory sets with $D_{\text{IL}}$ and continue fine-tuning $\pi_{\theta_{j-1}}$ by maximizing the following objective:
\begin{multline}
    \label{equation:obj-ev}
    \mathcal{J}_{\text{SI}}^j(\theta) = \mathbb{E}_{(q,\tau) \sim D_{\text{SI}} } \sum_{t=1}^T  \Big[ \log \pi_{\theta}( \\ a_t | q,  c_t^{\text{clip}'}, h_t) + \log \pi_{\theta}(h_t | q, c_t^{\text{clip}'})\Big],
\end{multline}
where $j=1,...,m$ denotes the times of optimization, $D_{\text{SI}} = D_{\text{IL}} \cup D_{ev}^j$ denotes the mixed trajectory set and $\pi_{\theta_{0}} $ is set to $ \pi_{\theta_{b}}$. 
The complete procedure is shown in Algorithm \ref{algo:webvoyager2} in Appendix \ref{sec:appendix-algo}.

\begin{table*}[t]
\vspace{0.1in}
\centering
\setlength{\tabcolsep}{1.6mm}{
\scalebox{0.88}{\begin{tabular}{@{}lcccccccc@{}}
\toprule \toprule
& Allrecipes & Amazon & Apple & ArXiv & GitHub & Booking & ESPN & Coursera \\ \midrule

%GPT-4o & \ \ 56.3\% & \ \  53.7\%  & \ \ 56.6\%  & \ \ 60.5\% & \ \ 57.7\% & \ \ 43.9\%  & \ \  44.0\%  & \ \  65.1\%  \\

OpenWebVoyager$_{\text{IL}}$& \ \ 17.8\% & \ \  12.2\%  & \ \ 20.9\%  & \ \ 14.0\% & \ \ 14.6\% & \ \ 9.1\%  & \ \  9.1\%  & \ \  31.0\%   \\

OpenWebVoyager$_{\text{iter-1}}$& \ \ 35.2\%  & \ \ 26.8\%   & \ \ 11.6\%  & \ \ 18.6\%  & \ \ 24.4\%  & \ \ 6.8\%  & \ \ 2.3\%   & \ \ 28.6\%   \\

OpenWebVoyager$_{\text{iter-2}}$& \ \ 22.2\%  & \ \  36.6\%  & \ \  27.9\% & \ \ 20.9\%  & \ \ 19.5\% & \ \  6.8\%  & \ \ 6.8\%   & \ \ 33.3\%   \\
OpenWebVoyager$_{\text{iter-3}}$& \ \ 24.4\% & \ \ 24.4\%    & \ \  20.9\% & \ \ 18.6\% & \ \ 31.7\% & \ \  18.2\% & \ \ 11.4\%   & \ \  42.9\%  \\

OpenWebVoyager$_{\text{iter-3-dgs}}$& \ \ 20.0\% & \ \ 31.7\%   & \ \ 18.6\%  & \ \ 23.3\% & \ \  24.4\% & \ \  13.6\% & \ \  25.0\%  & \ \ 42.9\%   \\

OpenWebVoyager$_{\text{iter-3-dgs-g}}$& \ \ 22.2\% & \ \ 29.3\%   & \ \ 32.6\%  & \ \ 20.9\% & \ \  26.8\% & \ \  11.4\% & \ \  11.4\%  & \ \ 42.9\%   \\

\midrule \midrule
& Cambridge & BBC & Google & Google & Google & \multirow{2}{*}{Huggingface} & Wolfram & \multirow{2}{*}{Overall}  \\ 
& Dictionary & News & Flights & Map & Search & & Alpha & \\ 
\midrule 

%GPT-4o & \ \ 82.2\% & \ \  54.8\% & \ \ 28.6\%  & \ \ 56.9\% & \ \ 63.6\% & \ \ 42.6\%  & \ \  65.2\%  & \ \  55.5\%  \\

OpenWebVoyager$_{\text{IL}}$& \ \ 37.2\% & \ \   9.5\% & \ \ 9.5\%  & \ \ 22.0\% & \ \ 44.2\% & \ \ 20.9\%  & \ \  26.1\%  & \ \  19.9\%   \\

OpenWebVoyager$_{\text{iter-1}}$& \ \ 25.6\% & \ \  9.5\%  & \ \ 19.0\%  & \ \ 26.8\% & \ \ 44.2\%  & \ \ 25.6\%  & \ \  32.6\%  & \ \ 22.6\%   \\

OpenWebVoyager$_{\text{iter-2}}$& \ \ 23.3\%  & \ \  14.3\%  & \ \  19.0\% & \ \ 22.0\%  & \ \ 41.9\%  & \ \ 11.6\%  & \ \  34.8\%  & \ \  22.7\%  \\
OpenWebVoyager$_{\text{iter-3}}$& \ \ 37.2\% & \ \ 11.9\%   & \ \ 11.9\%  & \ \ 26.8\% & \ \  39.5\% & \ \  30.2\% & \ \  37.0\%  & \ \ 25.8\%   \\

OpenWebVoyager$_{\text{iter-3-dgs}}$& \ \ 30.2\% & \ \ 11.9\%   & \ \ 21.4\%  & \ \ 22.0\% & \ \  39.5\% & \ \  23.3\% & \ \  34.8\%  & \ \ 25.5\%   \\
OpenWebVoyager$_{\text{iter-3-dgs-g}}$& \ \ 34.9\% & \ \ 14.3\%   & \ \ 21.4\%  & \ \ 29.3\% & \ \  44.2\% & \ \  32.6\% & \ \  37.0\%  & \ \ 27.4\%   \\

\bottomrule \bottomrule
\end{tabular}}}
\vspace{-0.05in}
\caption{Task success rate on WebVoyager test set (643 queries). All websites are seen during training. `IL', `iter-1', `iter-2', and `iter-3' represent agents after IL, 1st, 2nd, and 3rd optimization, respectively. `dgs' and `dgs-g' denote difficulty-guided sampling, i.e., sample more trajectories for webs with low sampling accuracy, the former by adding trajectories sampled by the agent itself and the latter by adding trajectories sampled by GPT-4o.}  % The results of GPT-4o are taken from WebVoyager \citep{he2024webvoyager} as a reference. 
\vspace{0.1in}
\label{tab:main_result_webv}
\end{table*}

\begin{table*}[t]
\centering
\scalebox{0.8}{
\begin{tabular}{@{}lcccc|cccc@{}}
\toprule
\multirow{2}{*}{Agents} & \multicolumn{4}{c|}{Mind2Web cross-task (unseen task)} & \multicolumn{4}{c}{Mind2Web cross-web (unseen web)} \\
 & Entertainment   & Shopping   & Travel  & Overall  & Entertainment  & Shopping  & Travel  & Overall  \\ \midrule
  OpenWebVoyager$_{\text{IL}}$   &     8.2\%     &  5.9\%   &     4.3\%    &    6.3\%      &     3.0\%            &    13.3\%       &    4.7\%     &   6.6\%       \\
   OpenWebVoyager$_{\text{iter-1}}$  &      12.2\%      &        0.0\%    &    4.3\%     &     7.1\%     &  6.1\%          &      6.7\%     &    9.3\%     &       7.5\%   \\
   OpenWebVoyager$_{\text{iter-2}}$ &     24.5\%  &    5.9\%       &   6.5\%       &    14.3\%     &    15.2\%       &     10.0\%     &    7.0\%      &    10.4\%      \\
  OpenWebVoyager$_{\text{iter-3}}$  &     26.5\%      &  23.5\%          &    10.9\%     &    19.6\%      &  6.1\%              &    20\%       &    7.0\%     & 10.4\%         \\ 
  OpenWebVoyager$_{\text{iter-3-dgs}}$& 18.4\%  & 23.5\%  & 10.9\%  &  16.1\% &  9.1\%
 &  16.7\%  &  25.6\%  &  17.9\% \\ 
 OpenWebVoyager$_{\text{iter-3-dgs-g}}$& 22.4\%  &  29.4\% & 15.2\%  & 20.5\%  & 3.0\%
 &  20.0\%  &  23.3\%  & 16.0\% \\  \bottomrule
\end{tabular}}
\vspace{-0.05in}
\caption{Task success rate on Mind2Web cross-task and cross-web test set.  In cross-task set, the queries from the same websites are seen during training. In cross-website set, the websites are not seen during training but still belong to the Entertainment, Shopping, and Travel Domain. }
\label{tab:mind2web_main_res}
\end{table*}

\section{Experiment}

\subsection{Dataset and Metric}

\paragraph{Training Dataset} In  \S\ref{section:IL_stage}, we have outlined the composition of the query set $Q_{\text{IL}}$ during the Imitation Learning stage, which includes 48 websites mentioned in Mind2Web \citep{deng2024mind2web} and WebVoyager \citep{he2024webvoyager}, along with 1516 relevant task queries collected. We use WebVoyager-4o to gather corresponding trajectories for them, with each query having a maximum of 2 trajectories. Then we retain 1165 \textbf{finished} (including both successful and unsuccessful) trajectories, with a total of 7253 interaction turns. During the $j$-th exploration-feedback-optimization cycle, we expend 480 queries for 48 selected websites. The trajectories are sampled via $\pi_{\theta_{j-1}}$ and the maximum resampling count is set to 5. 

\paragraph{Evaluation Dataset} To evaluate the performance of our agent, we use the following datasets: 1) WebVoyager \citep{he2024webvoyager} test set, comprising 15 websites seen during training and 643 task queries; 2) Mind2Web \citep{deng2024mind2web} cross-task test set, which included 33 websites seen during training and a total of 112 queries. 3) Mind2Web cross-website test set, we select 2 websites each from the ``Entertainment'', ``Shopping'', and ``Travel'' domains, the websites are unseen during training but they have the same domains, amounting to a total of 106 queries. % 4) Mind2Web cross-domain test set, websites belong to ``Info'' and ``Service'' domains that are not seen during training, from which we select 6 domains and 12 websites, totaling 216 queries. 
\paragraph{Metric}
Following WebVoyager, we adopt Task Success Rate automatically evaluated by GPT-4o as the primary metric. To view the exploration efficiency in the exploration-feedback-optimization cycle, we define Success@K (S@K) as the ratio of tasks that get success within K samples. Additionally, we pay attention to the finish rate (F@1), where a task is considered finished as long as the agent selects `ANSWER' within the maximum navigation steps. Table \ref{tab:query_traj_details} shows the details of the query set and collected trajectories in exploration-feedback-optimization cycles.

\subsection{Experimental Details}
To collect data for imitation learning phase, we adopt the state-of-the-art model GPT-4o with WebVoyager framework (WebVoyager-4o) to %as an expert and 
sample web navigation trajectories. We set $k=3$, i.e., the context contains at most 3 screenshots and corresponding accessibility trees but retains the thoughts and actions generated by GPT-4o in each step. Our agent builds upon Idefics2-8b-instruct with outstanding vision-language capabilities to complete the imitation learning and exploration-feedback-optimization cycles. During fine-tuning, the max sequence length is set to 8192. We no longer use system prompts and further clip the context to accept a maximum of 3 screenshots and 1 accessibility tree. 
The original resolution of the screenshots is 1024*768 and the screenshots are resized such that the longer length is no larger than 980, before feeding into Idefics2.
% but in Idefics2, the screenshots are scaled down, with the maximum length set to 980. 
We set the batch size to 64 and train for 300 iterations in each phase, approximately 2 - 3 epochs. In the exploration-feedback-optimization phase, we iteratively train our agent with a total of $m=3$ iterations. When the agent performs exploration, we set the temperature to 1.2 to improve the randomness. The agent samples up to 5 trajectories for each given task query. We still select GPT-4o as the feedback model and trajectories with positive feedback are gathered for further optimizations.

\begin{figure}[t!]
\centering
\includegraphics[width=0.99\linewidth]{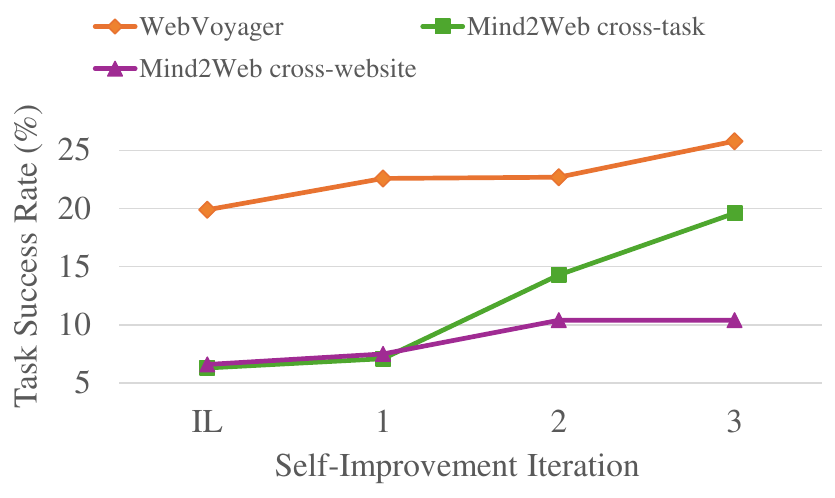}
\vspace{-0.2in}
\caption{Performance growth of OpenWebVoyager on WebVoyager and Mind2Web test set from Imitation Learning phase to 3rd exploration-feedback-optimization cycle.}
\vspace{0.1in}
\label{fig:improvement_curve}
\end{figure}

\begin{table*}[t]
\centering
\scalebox{0.9}{
\begin{tabular}{@{}ccccccccccc@{}}
\toprule
\begin{tabular}[c]{@{}c@{}} Improvement\\ Iteration\end{tabular}              & Query        & \begin{tabular}[c]{@{}c@{}} Traj.\\ From\end{tabular}   & \begin{tabular}[c]{@{}c@{}} Success\\ Traj.\end{tabular}   & Turns  & F@1 & S@1 & S@2 & S@3 & S@4 & S@5 \\ \midrule
iter-1   & 480 & $\pi_{\theta_b}$ &  152    &   943     &      74.6\%    &   10.4\%    &     19.6\%      &    24.4\%    &  27.5\%    &   31.7\%   \\
iter-2   & 480 & $\pi_{\theta_1}$ &  205    &   1324   &    87.1\%     &     16.0\%      &    24.0\%       &     30.2\%        &      36.9\%    &    42.7\%       \\
iter-3   & 480 & $\pi_{\theta_2}$ &  207    &   1333   &        91.5\%     &   18.8\%        &    27.9\%       &   35.2\%       &      41.0\%     &     43.1\%     \\ \bottomrule
\end{tabular}}
\vspace{-0.05in}
\caption{Details of query set and trajectory set during the exploration-feedback-optimization cycle. The feedback on task success or not is provided by GPT-4o. F@1 indicates the finish rate of the first exploration. S@K represents the task success rate within K explorations.
Each task will sample the trajectory up to 5 times until it succeeds or fails all 5 times, successful trajectories will be retained to improve our agent.}%thus S@5 can also represent the utilization of queries.}
\label{tab:query_traj_details}
\end{table*}

% \begin{table*}[t]
% \centering
% \scalebox{0.8}{
% \begin{tabular}{@{}lcccccccc@{}}
% \toprule
% \multirow{2}{*}{Agent} & \multicolumn{2}{c}{WebVoyager} & \multicolumn{2}{c}{\begin{tabular}[c]{@{}c@{}}Mind2Web \\ cross-task\end{tabular}} & \multicolumn{2}{c}{\begin{tabular}[c]{@{}c@{}}Mind2Web \\ cross-website\end{tabular}} & \multicolumn{2}{c}{\begin{tabular}[c]{@{}c@{}}Mind2Web\\ cross-domain\end{tabular}} \\
%  & Finish & Success & Finish & Success & Finish & Success & Finish & Success \\ \midrule
% OpenWebVoyager$_{\text{IL}}$ & 6.47 & 5.26 & 8.77 & 7.00 & 9.28 & 9.29 & 7.61 & 6.48 \\
% OpenWebVoyager$_{\text{iter-1}}$ & 6.17 & 5.02 & 7.58 & 5.00 & 7.98 & 9.63 & 6.92 & 9.27 \\
% OpenWebVoyager$_{\text{iter-2}}$ & 5.89 & 5.04 & 7.33 & 6.31 & 7.13 & 7.45 & 6.50 & 7.23 \\
% OpenWebVoyager$_{\text{iter-3}}$ & 5.47 & 5.07 & 7.67 & 7.59 & 6.16 & 6.91 & 5.68 & 7.15 \\ \bottomrule
% \end{tabular}}
% \caption{The average length of trajectories across different improving cycles on various test sets.   `Finish' and `Success' indicates that we calculate the average length for finished or successful trajectories, respectively.}
% \label{tab:traj_length}
% \end{table*}

\begin{table}[t]
\centering
\scalebox{0.7}{
\setlength{\tabcolsep}{0.8mm}{
\begin{tabular}{@{}lccccccc@{}}
\toprule
\multirow{2}{*}{Agent} & \multicolumn{2}{c}{WebVoyager} & \multicolumn{2}{c}{\begin{tabular}[c]{@{}c@{}}Mind2Web \\ cross-task\end{tabular}} & \multicolumn{2}{c}{\begin{tabular}[c]{@{}c@{}}Mind2Web \\ cross-website\end{tabular}}  \\
 & Finish & Success & Finish & Success & Finish & Success \\ \midrule
OpenWebVoyager$_{\text{IL}}$ & 6.47 & 5.26 & 8.77 & 7.00 & 9.28 & 9.29  \\
OpenWebVoyager$_{\text{iter-1}}$ & 6.17 & 5.02 & 7.58 & 5.00 & 7.98 & 9.63  \\
OpenWebVoyager$_{\text{iter-2}}$ & 5.89 & 5.04 & 7.33 & 6.31 & 7.13 & 7.45  \\
OpenWebVoyager$_{\text{iter-3}}$ & 5.47 & 5.07 & 7.67 & 7.59 & 6.16 & 6.91  \\ \bottomrule
\end{tabular}}}
\vspace{-0.05in}
\caption{The average length of trajectories across different optimization cycles on various test sets. `Finish' and `Success' indicates that we calculate the average length for finished or successful trajectories, respectively.}
\label{tab:traj_length}
\end{table}

\begin{table}[t]
\centering
\scalebox{0.8}{
\begin{tabular}{@{}lccccccccc@{}}
\toprule
\multirow{2}{*}{Agent} & \multicolumn{5}{c}{WebVoyager (643 tasks)}  \\
 & R & RS & S & RS / R & RS / S  \\ \midrule
 OpenWebVoyager$_{\text{IL}}$ & 61 & 8 & 128 &13.1\%  & 6.3\%  \\
 OpenWebVoyager$_{\text{iter-1}}$ & 75 & 16 & 145 & 21.3\% & 11.0\%  \\
 OpenWebVoyager$_{\text{iter-2}}$& 88 &  22 & 146 & 25.0\%&  15.1\% \\
 OpenWebVoyager$_{\text{iter-3}}$& 142 & 40 & 166 & 28.2\%&  24.1\% \\ \bottomrule
\end{tabular}}
\vspace{-0.05in}
\caption{The frequency of the agent using the restart action: Let R denote the number of trajectories with restart, RS the number of successful trajectories with restart, and S the total number of successful trajectories.}
\label{tab:restart}
\end{table}

\subsection{Main Results}
Throughout the entire process of Imitation Learning and exploration-feedback-optimization cycles, we trained four models: OpenWebVoyager$_{\text{IL}}$, OpenWebVoyager$_{\text{iter-1}}$, OpenWebVoyager$_{\text{iter-2}}$, and OpenWebVoyager$_{\text{iter-3}}$. Table \ref{tab:main_result_webv} shows the performance of these models on the WebVoyager test set. Table \ref{tab:mind2web_main_res} presents the results of these models on the Mind2Web cross-task and cross-website test set. % To further evaluate the models' out-of-domain (OOD) capabilities, we test these models on Mind2Web cross-domain test set. Table \ref{tab:mind2web_cross_domain} showcases the cross-domain improvement of the aforementioned models. 
We show the performance changes of our agent on these datasets from imitation learning phase to the third exploration-feedback-optimization cycle in Figure \ref{fig:improvement_curve}.

From the results in Table \ref{tab:main_result_webv} and Table \ref{tab:mind2web_main_res},  we observe a general improvement in task success rates in both the WebVoyager test set and the Mind2Web cross-task test set as optimization progressed. This indicates the effectiveness of our method when the webs in the test set have been trained on or explored during the training phase. In the Mind2Web cross-web test set, the exploration-feedback-optimization cycle also provides some enhancement in the model's performance, although not as prominently as in the cross-task set. Also, the improvement is unstable on these unexplored websites, agent suffers from sampling randomness 
and are more likely to get stuck during web navigation. 

%The \textbf{cross-domain results} in Table \ref{tab:mind2web_cross_domain} demonstrate that in cross-domain situations, the self-improving cycle provides minimal enhancement. This might be attributed to the agent's insufficient basic and generalizable abilities in web navigation. We notice that the agent often struggled to locate the elements requiring interaction in trajectories, leading to being stuck on a particular webpage. This emphasizes the importance of exploring unknown web pages. While the number of web pages is infinite, we believe that a sufficient amount of exploration-feedback-improve cycles can facilitate the emergence of the agent's generalization capacity for out-of-domain scenarios. 

Table \ref{tab:query_traj_details} shows the results of GPT-4o's feedback on the trajectories sampled by the agent during the exploration phase. We find that despite having 5 chances for resampling, 
% many websites still exhibit low accuracy. 
The agent still performs poorly on many websites.
Therefore, we consider increasing the number of trajectories specifically for these ``difficult'' websites during exploration-feedback-optimization phase.  To investigate the effectiveness of this \textbf{difficulty-guided sampling (DGS)} strategy, we train OpenWebVoyager$_{\text{iter-3-dgs-g}}$ and OpenWebVoyager$_{\text{iter-3-dgs}}$. The former involves adding some trajectories sampled by WebVoyager-4o for webs with S@5 below 40\% during the third iteration, while the latter adds some trajectories sampled by the agent itself. Compared to OpenWebVoyager$_{\text{iter-3}}$, adding exploration trajectories to the ``difficult'' websites can improve performance for certain websites like Google Flights. However, influenced by the sampling randomness, the optimization is not stable, as seen in Booking, GitHub, and others. Additionally, incorporating WebVoyager-4o sampled trajectories during the exploration phase has resulted in some overall performance enhancements.

\subsection{Discussion}
\paragraph{The average length of trajectories.}
During inference, we record the length of trajectories when they are finished (the agent provides answers) and successful. The variation of the average length of web navigation trajectories is shown in Table \ref{tab:traj_length}. 

In our experiments, we observe that as iterative optimization progresses, agents tend to complete tasks in fewer interaction steps and navigate more quickly on familiar websites. This phenomenon creates a cycle where trajectories obtained during the exploration-feedback phase become shorter, leading the model to increase its focus on learning from shorter trajectories during optimization.

\paragraph{Hallucination limits agent's performance.}
We find that agents often directly hallucinate answers that do not appear during the navigation process. The decrease in trajectory length might have increased the frequency of this issue. The agent tends to terminate navigation directly instead of continuing the search after a certain length of the trajectory. As shown in Table \ref{tab:query_traj_details}, we can also observe that the results for F@1 are high, but S@1 are relatively low. This indicates that the agent believes it has finished the task but is actually unsuccessful. While the finish rate and success rate in GPT-4o-sampled trajectories are close. This insight suggests that in future exploration, we can increase the diversity of sampling by varying the task difficulty and trajectory length.

\paragraph{Restart to the search engine and solve tasks.}
In WebVoyager's paradigm, an important web action is to restart navigation from the search engine when encountering difficulties. In this paper, the `Restart' action is also provided in the data for training during the Imitation Learning phase. 
We observe the frequency of our agent using restart action, calculate their success rates, and the ratio of successful tasks using restart to the total successful tasks, as shown in Table \ref{tab:restart}. We can infer from the results in the WebVoyager test set that as agents undergo iterative optimization, they increasingly prefer to use the search engine. The proportion of successful trajectories achieved by using the search engine is rising among all successful trajectories, addressing some of the navigation failure issues.

\begin{table}[t]
\centering
\scalebox{0.8}{
\begin{tabular}{@{}cc@{}}
\toprule
 \begin{tabular}[c]{@{}c@{}}Training\\ Trajectories\end{tabular} & Result \\ \midrule
 $D_{\text{IL}} \cup D_{\text{iter-1}} \cup D_{\text{iter-2}}$ & 20.8\% \\
 $D_{\text{IL}} \cup D_{\text{iter-2}}$ & 23.3\% \\ \bottomrule
\end{tabular}}
\vspace{-0.05in}
\caption{Study on whether to use a mixture of data from previous phases in exploration-feedback-optimization cycle (OpenWebVoyager$_{\text{iter-1}}$ $\rightarrow$ OpenWebVoyager$_{\text{iter-2}}$). }
\label{tab:used_trajectories}
\end{table}

\paragraph{Other settings and parameters.}
Trajectory collection is time-consuming, especially in the exploration phase where each query requires up to 5 resampled trajectories to tackle relatively difficult navigation tasks. So we primarily adjust hyperparameters such as learning rate and global batch size during the IL phase. However, we ultimately find that this has little significance, as the error is much smaller compared to the challenges posed by webpage navigation and the sampling randomness. 

In exploration-feedback-optimization cycles, we also attempt to mix all trajectories that considered success through GPT-4o's feedback, for example, using $D_{\text{IL}} \cup D_{\text{iter-1}} \cup D_{\text{iter-2}}$ to improve OpenWebVoyager$_{\text{iter-1}}$. We select 120 WebVoyager queries and compare task success rate in Table \ref{tab:used_trajectories}. % and obtain OpenWebVoyager$_{\text{iter-2}}'$. We select 120 WebVoyager queries and find that the task success rate of OpenWebVoyager$_{\text{iter-2}'}$ is \textbf{20.8\%}, lower than OpenWebVoyager$_{\text{iter-2}}$ obtained by solely using $D_{\text{IL}} \cup D_{\text{iter-2}}$, which is \textbf{23.3\%}.

\section{Conclusion}
In this paper, we explore how to construct a multimodal web agent via iterative exploration, feedback and optimization. We adopt idefics2-8b-instruct as the backbone LMM model and collect web task queries from numerous websites. Initially, our agent learns the web operation logic of GPT-4o through Imitation Learning. Then it enters the exploration-feedback-optimization cycles, exploring and collecting trajectories based on new web tasks, retaining the trajectories that GPT-4o considers correct for further learning, updating, and optimization. We focus on building an LMM-based iterative optimization web agent with multi-image understanding capabilities, enabling it to adapt to complex and dynamic online web environments. The entire process primarily involves the agent's self-exploration and GPT-4o's supervision, reducing human intervention and allowing continuous expansion to ensure the agent's generality.

\section*{Limitations}
%In our work, we primarily rely on the WebVoyager paradigm for web navigation.
First, we only consider the most common executable web actions in the simulated environment, including clicking, typing, and scrolling, without more advanced actions such as dragging and zooming. 
%Secondly, the web agent fine-tuned under this paradigm adapts to its operational mode, and using other system prompts to adjust the agent's generation may result in huge performance degradation.
%Additionally, the complex and dynamic online environment along with the relatively small parameter size of Idefics2 (only 8B), constrain the agent's performance. 
Additionally, our approach is based on a relatively small LMM Idefics2 with 8B parameters, which may limit the agent's ability to effectively navigate websites of unseen domains and respond to complex user queries.
The low performance on complex websites might further affect exploration efficiency, leading to minimal improvement and time-consuming during the exploration-feedback-optimization process.
Last, our model still primarily relies on accessibility trees, we hope to improve the visual grounding and multi-image reasoning capabilities so that it can directly use web screenshots for planning like GPT-4o.

\section*{Ethics Statement}
In light of the potential risks associated with online web navigation, all our experiments adhere strictly to ethical guidelines. Our approach includes human supervision as well as GPT-4's monitoring for content violations. Throughout the sampling of all web task trajectories, no violations by the agent are detected. A small portion of tasks are filtered due to the sensitivity of advertisements or content on news websites. None of the tasks involve private information such as personal names, account passwords, etc. The tasks typically include information-seeking activities and do not include actual bookings or payment transactions. In our work, the web agent's sampled trajectories are intended solely for research purposes. The agent operates in a simulated human-like manner, with a slow sampling frequency, ensuring no pressure is placed on the explored websites.

%\paragraph{Difficulty of websites affects self-improvements.}

% \section*{Acknowledgments}

\bibliography{custom}
\clearpage
\appendix

\section{Environment and Prompts}
\label{sec:appendix-env-prompt}
\begin{figure*}[t]
\centering
\includegraphics[width=0.85\linewidth]{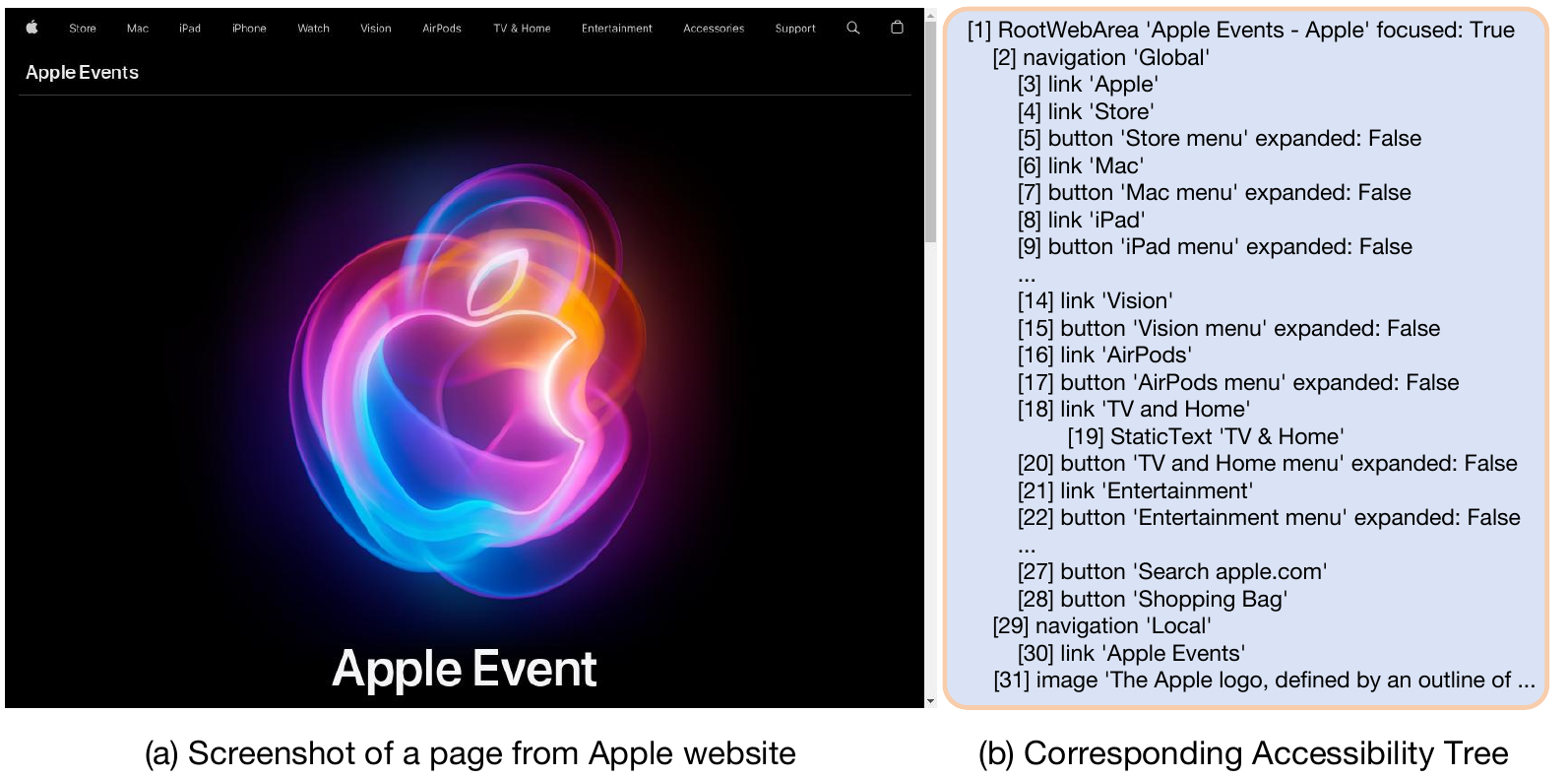}
\caption{An example of observations fed into the agent, where the screenshot is rendered by the browser, and the accessibility tree is extracted from the HTML and numbered starting from `[1]'.}
\label{fig:observation_space}
\vspace{-0.05in}
\end{figure*}

We adopt the framework of WebVoyager for online real-world web navigation. The web actions used are the most basic clicks, inputs, and scroll operations as shown in Table \ref{tab:web_actions}. Unlike WebVoyager, we do not use the Set-of-Mark approach to label screenshots. Instead, we combine screenshots and the accessibility tree as observations for the agent to make decisions. Figure \ref{fig:observation_space} illustrates an example of observation.

Based on the changes in observations, we slightly modify the system prompt of WebVoyager \citep{he2024webvoyager} during the Imitation Learning phase to accommodate the paradigm of accessibility tree + screenshot. In terms of web operation implementation, each element in the accessibility tree has pre-saved attribute information, where `union\_bound' labels the position information of the element.% we utilize the positional information of web elements provided in the accessibility tree. 
We use Selenium to locate the element that appears in this position and then access it.

In the WebVoyager framework, in addition to the system prompt, the author has designed error reflection to ensure effectiveness. When a certain action fails, there will be a prompt saying: "\texttt{The action you have chosen cannot be executed. Please double-check if you have selected the correct element or used the correct action format. Then provide the revised Thought and Action.}" This prompt serves to remind the agent to correct errors. While training our own Agent, although we no longer use the system prompt, we still retain the error reflection mechanism.

\begin{table*}[t]
\centering
\begin{tabular}{@{}lcll@{}}

\toprule
Web Actions & Format  & Notes \\ \midrule
Click & Click [Label] & 
\begin{tabular}[c]{@{}l@{}}Perform a single Click operation \\ on an web element.\end{tabular}

\\ \midrule
Input & Type [Label]; [Content]   &  
\begin{tabular}[c]{@{}l@{}}Type something in the text box \\ and press enter.\end{tabular}

\\ \midrule
Scroll & Scroll [WINDOW or Label]; [up or down] & 
\begin{tabular}[c]{@{}l@{}}In some web pages where only a \\ partial area can be scrolled, agent \\ need to lock an element in that \\area first, otherwise scrolls are \\ performed on the whole page.\end{tabular}

\\  \midrule

Go back & GoBack   & 
Go back to previous page

\\ \midrule
Restart & Restart  & 
\begin{tabular}[c]{@{}l@{}}Restart from Google Search \\ and solve tasks.\end{tabular}

\\ \midrule
Wait & Wait &  Sleep 5 seconds  \\ \midrule
Answer & ANSWER; [content] &  Provide final answer.  \\ \bottomrule
\end{tabular}
\caption{Web Actions used in this paper. }
\label{tab:web_actions}
\end{table*}

\begin{algorithm}[t]
\small
    \caption{OpenWebVoyager} %with Imitation Learning and Exploration-Feedback
    \label{algo:webvoyager2}
    \begin{algorithmic}
        \State \textbf{Input:} LMM-based Agent $\pi_\theta$, GPT-4o Agent $\pi_{\theta_g}$, GPT-4o Evaluator $\mathcal{R}_{\theta_g}$, query set $Q_{\text{IL}}$ for Imitation Learning, $Q_{\text{SI}}^1$,..., $Q_{\text{SI}}^m$ for exploration-feedback-optimization stages. 
        \State \textbf{Output:} The fine-tuned Agent $\pi_{\theta_m}$
        \Procedure{Imitation Learning:}{}
            \State $D_{\text{IL}}=\big\{(q_i, \tau_i) | q_i \in Q_{\text{IL}}, \tau_i \sim \pi_{\theta_g}(\tau|I, q_i) \big\}_{i=1}^{|D_{\text{IL}}|}$;
            \State Maximize  $\mathcal{J}_{\text{IL}}(\theta)$ shown in Equation \ref{equation:J-bc} to get $\pi_{\theta_b}$;
        \EndProcedure
        
        \Procedure{Exploration-Feedback-Optimization:}{}
            \State $\pi_{\theta_0} \leftarrow \pi_{\theta_b}$; 
            \For{iteration $j=1,...,m$}
                \State Collect trajectories $D_{\text{SI}}^j$ with rejection sampling:
                \State $D_{\text{SI}}^j \leftarrow \{\}$;
                % \State $D_{ev}^j = \big\{(q_i, \tau_i^l) | q_i \in Q_{ev}^j, (\tau_i^1,..., \tau_i^l) \sim \pi_{\theta_{j-1}}(\tau|I, q_i), \mathcal{R}_{\theta_g}(\tau_i^l) = 1 \big\}_{i=1}^{|D_{ev}^j|}$;
                \For{$q \in Q_{\text{SI}}^j$}
                    \While{$l<$ max resampling count}
                    \State $\tau_l \sim \pi_{\theta_{j-1}}(\tau| q) $;
                    \If{$\mathcal{R}_{\theta_g}(\tau_i^l) = 1$}
                    \State $D_{\text{SI}}^j \leftarrow
                    D_{\text{SI}}^j \cup \{\tau_l\}$;
                    \State break;
                    \EndIf
                    \EndWhile
                \EndFor
                \State $D_{\text{SI}} \leftarrow D_{\text{IL}} \cup D_{\text{SI}}^j$;
                \State Maximize $\mathcal{J}_{\text{SI}}^j(\theta)$ shown in Equation \ref{equation:obj-ev} to get $\pi_{\theta_j}$;
            \EndFor
             
        \EndProcedure
    \end{algorithmic}
\end{algorithm}

\section{Algorithm}\label{sec:appendix-algo}
In Algorithm \ref{algo:webvoyager2}, we present the complete algorithm of OpenWebVoyager. It mainly consists of an Imitation Learning (IL) phase and multiple exploration-feedback-optimization cycles. In the IL phase, GPT-4o ($\pi_{\theta_g}$) serves as an expert to sample trajectories via WebVoyager framework, requiring a significant number of OpenAI API calls. In the exploration-feedback-optimization cycle, GPT-4o acts as an expert to evaluate trajectories, with only one API call needed for each trajectory. Hence, during the execution of the algorithm, there is a trade-off. On one hand, we aim to increase the sampling in the IL phase to enhance the model's capabilities and obtain a strong base model ($\pi_{\theta_b}$), which can improve exploration efficiency. However, if the improvement in the IL phase is not obvious, using additional GPT-4o calls for the IL phase might not be cost-effective. In such cases, letting the agent explore on its own with GPT-4o serving as auxiliary supervision might be more beneficial.

\section{Details of Datasets}
\label{sec:appendix-datasets}

\paragraph{Selected Websites}
In the Imitation Learning phase and exploration-feedback-optimization cycles, we collect task queries from 48 websites for exploration. We utilize all 15 webs from WebVoyager and 37 webs from Mind2Web, totaling 48 webs (with 4 duplicates). Table \ref{tab:train_datasets} displays the specific website names used during the training phase. During inference, we employ all task queries from the WebVoyager test set and select some task queries from the Mind2Web cross-task and cross-website test set%, and cross-domain test set, 
including both learned and unlearned websites. To facilitate testing, we update the time information of some tasks but do not change their task expressions. Table \ref{tab:test_datasets} presents detailed statistics about the test set.

\paragraph{Queries preparation for Imitation Learning} 
The learning effectiveness during the Imitation Learning phase is not only related to the expertise of GPT-4o but also to the richness of the task queries used. To diversify trajectories as much as possible during the Imitation Learning phase, we collect task queries from the following perspectives:
\begin{itemize}
    \item Queries from Mind2Web Training Data. We have chosen 37 available websites along with their corresponding queries, updating the date information for travel-related tasks, totaling 516 queries.
    \item Synthesised queries via self-instruct. Employing the self-instruct \citep{wang2022self} based method mentioned in WebVoyager (15 websites), we have generated 20 queries for each website, resulting in a total of 300 queries.
    The sentence-embedding model all-mpnet-base-v2\footnote{\url{https://huggingface.co/sentence-transformers/all-mpnet-base-v2}} is used to calculate the query similarity and filter out the queries with high similarity to ensure task diversity.
    There are 4 websites overlapping between WebVoyager and Mind2Web, making a total of 48 websites.
    \item Human-written queries. Recognizing the randomness and complexity of the above tasks, we borrow the idea of Curriculum Learning \citep{soviany2022curriculumlearningsurvey} and manually designed 5 easier task queries for each website, which can be completed by humans between 2 - 6 steps, amounting to a total of 240 tasks.
    \item General queries from users. To enhance generalization, we gather 460 queries provided by \citet{zhang2024cognitive}, and standardize them to begin navigation from search engines. This approach allows the agent to explore a wider range of websites and helps it recognize that in case of navigation failures, using a search engine could be attempted.
\end{itemize}

\begin{table*}[t]
\begin{tabular}{@{}lccc@{}}
\toprule
From & Domain & Subdomain & Website Name \\ \midrule
WebVoyager & - & - & \begin{tabular}[c]{@{}c@{}}Allrecipes; Amazon; Apple; ArXiv;\\ BBC News; Booking; Cambridge Dictionary;\\ Coursera; ESPN;GitHub; Google Flights;\\ Google Map; Google Search; Huggingface; Wolfram Alpha\end{tabular} \\ \midrule
\multirow{16}{*}{Mind2Web} & \multirow{5}{*}{Entertainment} & Event & eventbrite; nyc; ticketcenter \\
 &  & Game & boardgamegeek; store.steampowered \\
 &  & Movie & imdb; rottentomatoes; tvguide \\
 &  & Music & discogs; last.fm; soundcloud; \\
 &  & Sports & espn; foxsports; sports.yahoo; \\ \cmidrule(l){2-4} 
 & \multirow{4}{*}{Shopping} & Digital & apple \\
 &  & Fashion & uniqlo \\
 &  & General & amazon; ebay; target \\
 &  & Speciality & cvs; ikea \\ \cmidrule(l){2-4} 
 & \multirow{7}{*}{Travel} & Airlines & ryanair \\
 &  & Car rental & enterprise \\
 &  & General & agoda; booking \\
 &  & Ground & amtrak; mbta; thetrainline; us.megabus \\
 &  & Hotel & airbnb; koa; marriott \\
 &  & Restaurant & resy; yelp \\
 &  & Others & flightaware; nps.gov; spothero \\ \bottomrule
\end{tabular}
\caption{In the Imitation Learning and exploration-feedback-optimization cycles, a total of 48 websites are selected, including 15 from WebVoyager and 37 from Mind2Web (4 duplicates).}
\label{tab:train_datasets}
\end{table*}

\begin{table*}[t]
\scalebox{0.9}{
\begin{tabular}{@{}lccccc@{}}
\toprule
Test set & \begin{tabular}[c]{@{}c@{}}Num of \\ queries\end{tabular} & \begin{tabular}[c]{@{}c@{}}Web seen \\ in training?\end{tabular} & Domain & Subdomain & \begin{tabular}[c]{@{}c@{}}Websites and \\ num of queries\end{tabular} \\ \midrule
WebVoyager & 643 & Yes & - & - & \begin{tabular}[c]{@{}c@{}}Allrecipes: 45; Amazon: 41; Apple: 43; \\ ArXiv: 43; BBC News: 42; Booking: 44; \\ Cambridge Dictionary: 43; Coursera: 42; \\ ESPN: 44; GitHub: 41; Google Flights: 42;\\ Google Map: 41; Google Search: 43; \\ Huggingface: 43; Wolfram Alpha: 46\end{tabular} \\ \midrule
\multirow{15}{*}{\begin{tabular}[c]{@{}l@{}}Mind2Web \\ cross-task\end{tabular}} & \multirow{15}{*}{112} & \multirow{15}{*}{Yes} & \multirow{5}{*}{Entertainment} & Event & eventbrite: 6; nyc: 3; ticketcenter: 4 \\
 &  &  &  & Game & boardgamegeek: 1; store.steampowered: 1 \\
 &  &  &  & Movie & imdb: 5; rottentomatoes: 1; tvguide: 3 \\
 &  &  &  & Music & discogs: 6; last.fm: 5; soundcloud: 4 \\
 &  &  &  & Sports & espn: 4; foxsports: 5; sports.yahoo: 1 \\ \cmidrule(l){4-6} 
 &  &  & \multirow{4}{*}{Shopping} & Digital & apple: 4 \\
 &  &  &  & Fashion & uniqlo: 3 \\
 &  &  &  & General & amazon: 2; target: 5 \\
 &  &  &  & Speciality & cvs: 1; ikea: 2 \\ \cmidrule(l){4-6} 
 &  &  & \multirow{6}{*}{Travel} & Airlines & ryanair: 6 \\
 &  &  &  & General & agoda: 3; booking: 2 \\
 &  &  &  & Ground & amtrak: 6; mbta: 4; us.megabus: 1 \\
 &  &  &  & Hotel & airbnb: 3; koa: 3; marriott: 5 \\
 &  &  &  & Restaurant & resy: 2; yelp: 4 \\
 &  &  &  & Other & flightaware: 4; spothero: 3 \\ \midrule
\multirow{6}{*}{\begin{tabular}[c]{@{}l@{}}Mind2Web \\ cross-website\end{tabular}} & \multirow{6}{*}{106} & \multirow{6}{*}{No} & \multirow{2}{*}{Entertainment} & Event & stubhub: 16 \\
 &  &  &  & Sports & nba: 17 \\ \cmidrule(l){4-6} 
 &  &  & \multirow{2}{*}{Shopping} & Auto & cars: 13 \\
 &  &  &  & General & shopping.google: 17 \\ \cmidrule(l){4-6} 
 &  &  & \multirow{2}{*}{Travel} & Restaurant & tripadvisor: 23 \\
 &  &  &  & Other & recreation.gov: 20 \\ \bottomrule
 %\midrule
% \multirow{6}{*}{\begin{tabular}[c]{@{}l@{}}Mind2Web \\ cross-domain\end{tabular}} & \multirow{6}{*}{216} & \multirow{6}{*}{No} & \multirow{3}{*}{Info} & Finance & coinmarketcap: 21; finance.google: 20 \\
%  &  &  &  & Housing & student: 22; redfin: 17 \\
%  &  &  &  & Weather & weather: 9; theweathernetwork: 16 \\ \cmidrule(l){4-6} 
%  &  &  & \multirow{3}{*}{Service} & Government & ca.gov: 15; gov.uk: 17 \\
%  &  &  &  & Health & drugs: 23; healthgrades: 22 \\
%  &  &  &  & Pet & akc.org: 19; petfinder: 15 \\ \bottomrule
\end{tabular}}
\caption{Detailed statistics of the test dataset. Websites from WebVoyager and Mind2Web cross-task have been seen during training, while websites from Mind2Web cross-websites have not been encountered.}
\label{tab:test_datasets}
\end{table*}

\begin{figure*}[t]
\centering
\includegraphics[width=1.01\linewidth]{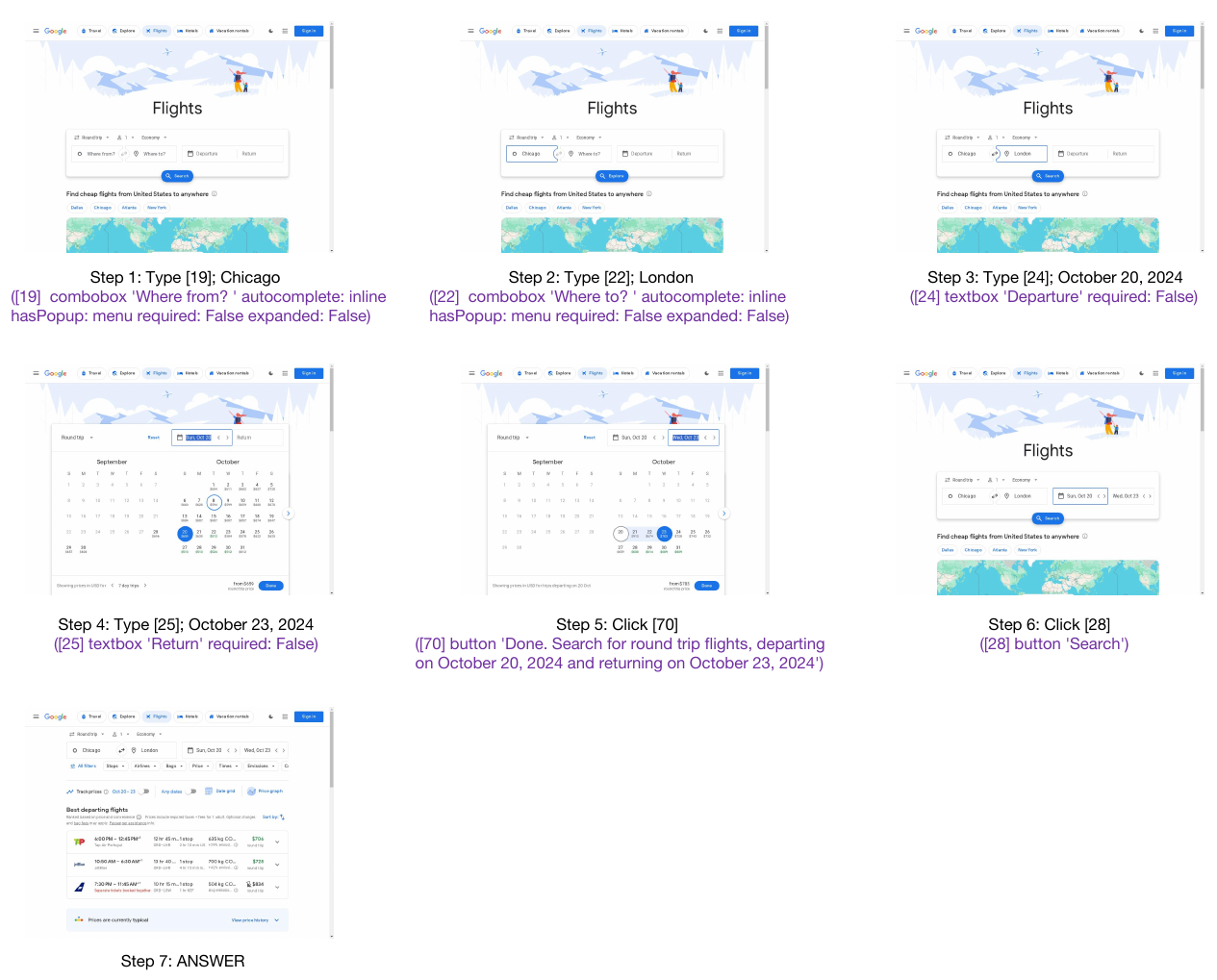}
\caption{Agent (OpenWebVoyager$_{\text{iter-3}}$) navigates on the Google Flights and completes the task successfully.
\textbf{TASK}: "\textit{Find flights from Chicago to London on 20 October and return on 23 October.}" \textbf{REPLY}: "\textit{Thought: The search results for flights from Chicago to London on October 20 and October 23, 2024, are displayed. The prices start from \$706 and go up to \$834. The average price for the round trip is approximately \$750. Action: ANSWER; The average price for a round trip flight from Chicago to London on October 20 and returning on October 23 is approximately \$750.}"}
\label{fig:google_flights5}
\vspace{-0.05in}
\end{figure*}

\begin{figure*}[t]
\centering
\includegraphics[width=1.01\linewidth]{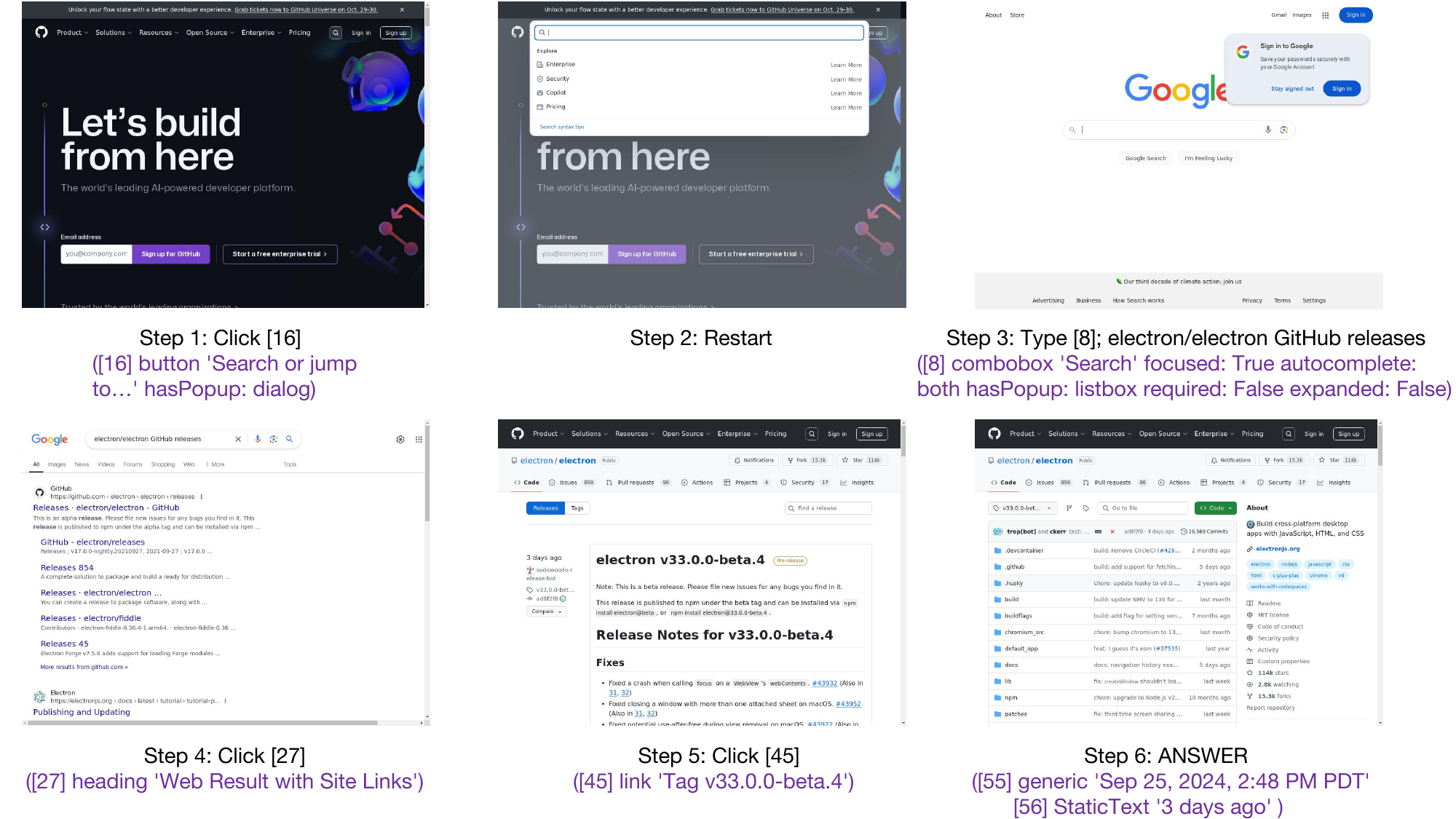}
\caption{Agent (OpenWebVoyager$_{\text{iter-3}}$) \textbf{Restart} navigation from Google Search with success after failing to navigate on GitHub pages. 
\textbf{TASK}: "\textit{Retrieve the latest release from the 'electron/electron' repository on GitHub and note down the release version number and date.}" \textbf{REPLY}: "\textit{Thought: I have navigated to the latest tag for the 'electron/electron' repository. The version number shows 33.0.0-beta.4, and the release date is Sep 25, 2024, as displayed on the right column. Action: ANSWER; Latest release of the 'electron/electron' repository is 33.0.0-beta.4 on Sep 25, 2024.}"}
\label{fig:github12}
\vspace{-0.05in}
\end{figure*}

\begin{figure*}[t]
\centering
\includegraphics[width=1.01\linewidth]{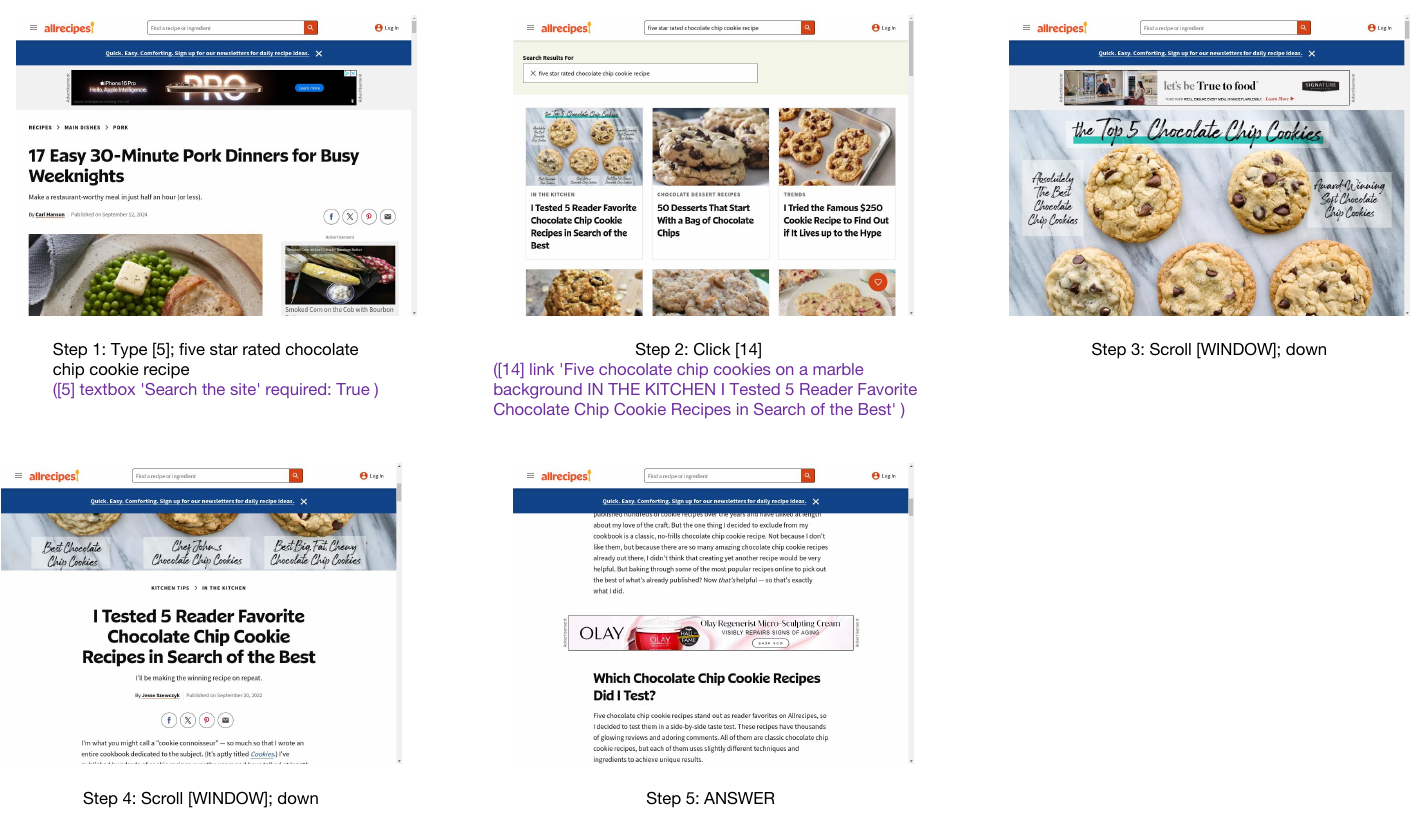}
\caption{Agent (OpenWebVoyager$_{\text{iter-3}}$) hallucinates an incorrect answer while navigating on Allrecipes. \textbf{TASK}: "\textit{Find a five-star rated chocolate chip cookie recipe that takes less than 1 hour to make on Allrecipes.}" \textbf{REPLY}: "\textit{Thought: I have found the section listing the main ingredients and the bake time for the "Classic Chocolate Chip Cookies." The cookies take 15 minutes to make and additional 45 minutes to bake, totaling 1 hour and 10 minutes. I now have all the necessary information to answer the task. Action: ANSWER; The \"Classic Chocolate Chip Cookies\" recipe on Allrecipes has a rating of 4.5 stars with 4,580 reviews. The main ingredients include butter, heavy cream, eggs, all purpose flour, and unsweetened cocoa powder. The cookies take 15 minutes to make and additional 45 minutes to bake, totaling 1 hour and 10 minutes.}"}
\label{fig:allrecipes_16}
\vspace{-0.05in}
\end{figure*}

\section{Example Trajectories}
In Figures \ref{fig:google_flights5} and \ref{fig:github12}, we present two examples of successful webpage navigations by OpenWebVoyager$_{\text{iter-3}}$. As shown in Figure \ref{fig:google_flights5}, agent navigates directly on the Google Flights webpage and succeeds. The agent makes decisions based on the screenshots and the specific text information of web elements in the accessibility trees. In Figure \ref{fig:github12}, the agent mistakenly thinks that logging in is required to search on GitHub, then it chooses to restart from Google Search and finds the answer.

We also present an example where an agent hallucinates an answer when it cannot find one. As Illustrated in Figure \ref{fig:allrecipes_16}, while navigating the Allrecipes website, the agent fails to locate a chocolate chip cookie recipe that meet the task requirements. However, it provides an answer titled "Classic Chocolate Chip Cookies." This discrepancy may be attributed to the agent interpreting the word "Classic" in the accessibility trees as a recipe and even hallucinating a cook time, despite the lack of relevance.

\end{document}